\documentclass[twocolumn, switch]{article}

\usepackage{amsmath}
\usepackage{amssymb}
\usepackage{preprint}
\usepackage[numbers,square]{natbib}

\usepackage[utf8]{inputenc} 
\usepackage[T1]{fontenc}    
\usepackage{url}            
\usepackage{booktabs}       
\usepackage{amsfonts}       
\usepackage{nicefrac}       
\usepackage{microtype}      
\usepackage{graphicx}
\usepackage{xcolor}
\usepackage{caption}        
\usepackage{array}
\usepackage{algorithm}
\usepackage{algorithmic}
\usepackage[colorlinks=true,
            linkcolor=purple,
            urlcolor=cyan,
            citecolor=cyan,
            anchorcolor=black]{hyperref}

\newsavebox{\mainresultstablebox}

\title{From Likelihood to Utility: Quality-Axis Diffusion\\ for End-to-End Autonomous Driving}
\author{
  {\bfseries Junjie Yang\textsuperscript{1} \quad Ziyi Shi\textsuperscript{1} \quad Shuqi Shen\textsuperscript{2} \quad Xinhu Zheng\textsuperscript{3} \quad Hongliang Lu\textsuperscript{4} \quad Hai Yang\textsuperscript{1}}
}

\hypersetup{
pdftitle={From Likelihood to Utility: Quality-Axis Diffusion for End-to-End Autonomous Driving},
pdfsubject={cs.RO, cs.AI},
pdfauthor={Junjie Yang, Ziyi Shi, Shuqi Shen, Xinhu Zheng, Hongliang Lu, Hai Yang},
}

\begin{document}

\twocolumn[
  \begin{@twocolumnfalse}
\maketitle
\thispagestyle{empty}
\begin{abstract}
Diffusion models have become a dominant paradigm for end-to-end autonomous driving, yet existing planners organize trajectory generation along the Gaussian-noise axis, treating driving utility only as an auxiliary guidance or selection signal. We instead formulate diffusion along a driving-quality axis. We propose QualiDrive, which constructs scene-specific quality ladders by ranking observed trajectories according to measured driving utility and learns to reverse this ordered degradation through
ordered low-to-high-quality refinement. Each reverse step is conditioned on the scene, ego state, current trajectory, and target quality level, progressively refining trajectories toward higher driving quality. On NAVSIM, QualiDrive achieves 94.5 PDMS on v1 and a state-of-the-art 92.74 EPDMS on v2. QualiDrive improves Oracle EPDMS only modestly over endpoint-only quality supervision (98.52 to 98.57), but substantially increases the fractions of candidates above 0.95 and 0.98 EPDMS from
12.94\%/9.71\% to 21.53\%/15.93\%, raising selected EPDMS from 92.63 to
92.74. These results suggest that diffusion is not merely a process of denoising, but one of progressive quality-ordered refinement.
\end{abstract}
\vspace{0.35cm}
  \end{@twocolumnfalse}
]

\begingroup
\let\thefootnote\relax
\footnotetext{\hangindent=2.3em\hangafter=1\textsuperscript{1}\,The Hong Kong University of Science and Technology}
\footnotetext{\hangindent=2.3em\hangafter=1\textsuperscript{2}\,The Chinese University of Hong Kong, Shenzhen}
\footnotetext{\hangindent=2.3em\hangafter=1\textsuperscript{3}\,The Hong Kong University of Science and Technology (Guangzhou)}
\footnotetext{\hangindent=2.3em\hangafter=1\textsuperscript{4}\,Southern University of Science and Technology}
\endgroup

\section{Introduction}

\begin{figure}[!t]
    \centering
    \includegraphics[
        width=\linewidth
    ]{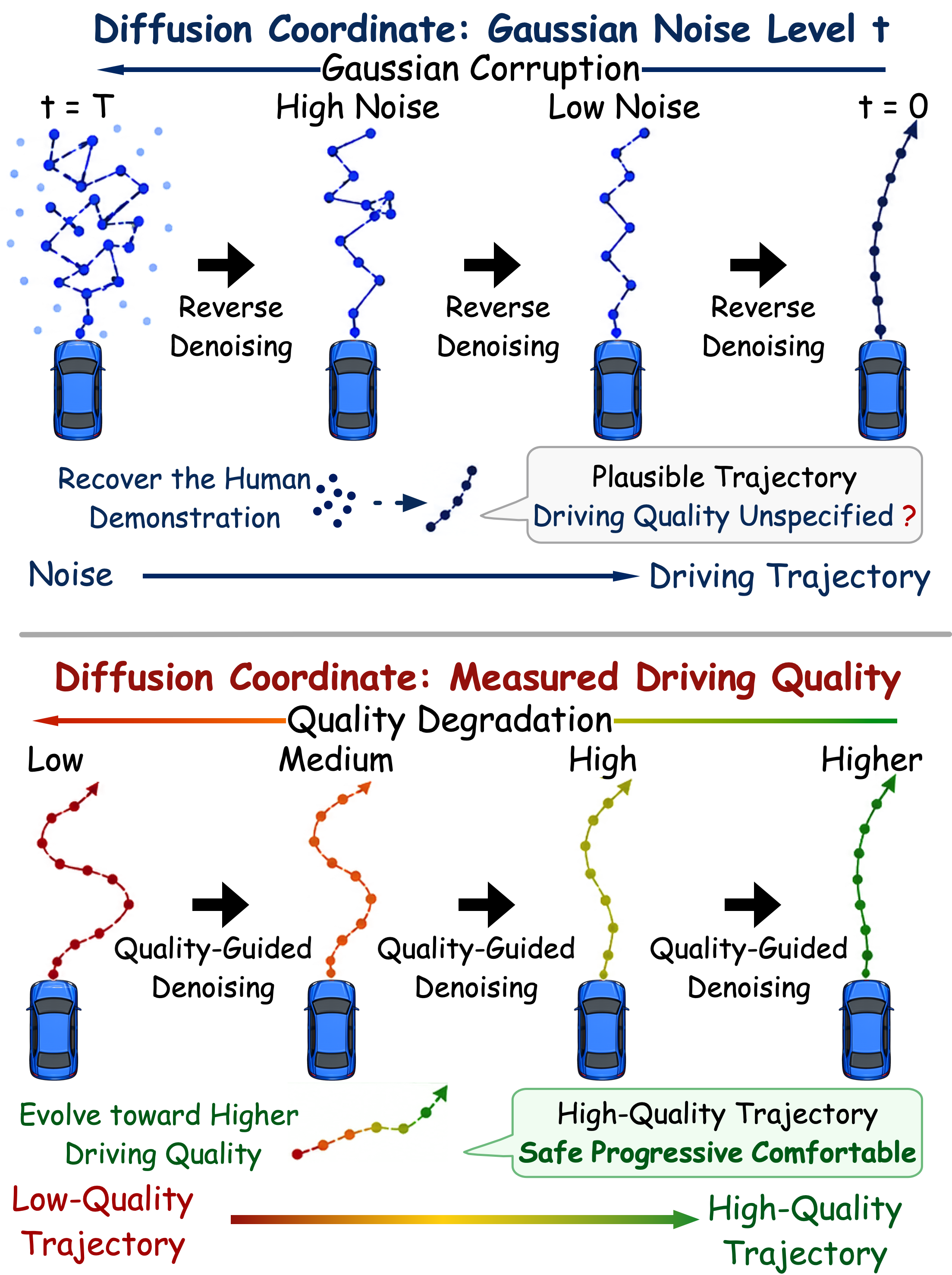}
    \caption{\textbf{Redefining the diffusion coordinate.}
    Upper: Gaussian-noise diffusion recovers plausible human demonstrations of
    unspecified quality. Lower: QualiDrive follows measured quality toward safer,
    more progressive, and more comfortable trajectories.}
    \label{fig:teaser}
\end{figure}

End-to-end (E2E) autonomous driving has emerged as a promising paradigm for directly  mapping sensor observations to future trajectories
~\citep{chitta2022transfuser,hu2022stp3,hu2023uniad,jiang2023vad,chen2024e2esurvey}. 
Leveraging large-scale human driving demonstrations, existing E2E systems primarily learn to generate trajectories with high likelihood under the human demonstration distribution ~\citep{cheng2024pluto,codevilla2018cil}.
However, high likelihood does not necessarily imply high driving utility, as human driving demonstrations are not necessarily optimal, and high-utility driving must simultaneously account for multiple objectives, such as safety, progress, and comfort~\citep {dauner2023pdmclosed,dauner2024navsim}. 

Recently, diffusion models have been increasingly adopted for trajectory planning due to their ability to generate diverse candidates ~\citep{liao2025diffusiondrive,zheng2025diffusionplanner,rombach2022ldm}. Conventional diffusion-based planners are organized along a Gaussian noise axis: their denoisers are trained to recover clean human demonstrations from Gaussian-corrupted trajectories, learning to eliminate synthetic perturbations. 
Yet diversity alone provides no guarantee of higher utility, and additional denoising steps may not translate into better plans~\citep{puthumanaillam2026muninn,song2023consistency}. 
This reveals a fundamental mismatch between the generative process and the planning objective: conventional diffusion progresses toward likely demonstrations, whereas planning ultimately seeks high-quality trajectories.


Prior work has sought to improve diffusion-based planning from both the generation and selection sides. Generation-oriented methods introduce reinforcement learning or safety guidance to steer diffusion toward higher-quality trajectories~\citep{zou2025diffusiondrivev2,zheng2025diffusionplanner}, while selection-oriented methods enhance candidate ranking through progressive scoring or multi-teacher distillation~\citep{yao2025drivesuprim,li2025hydra}. Despite using driving utility for guidance or selection, diffusion remains organized along the noise axis, with driving utility serving only as an auxiliary signal. Consequently, the generated candidate pool may still contain many low-quality trajectories, placing a heavy burden on downstream selection. This motivates a more fundamental question: can driving utility itself serve as the semantic coordinate of trajectory diffusion?

In this study, we propose QualiDrive, a quality-axis diffusion framework for E2E trajectory generation.
As illustrated in Figure~\ref{fig:teaser}, we organize the diffusion process by ordered quality levels rather than Gaussian-noise levels, transforming reverse diffusion from trajectory reconstruction into progressive quality refinement.
Specifically, the forward process constructs a sequence of quality-degradation transitions from a high-scoring target trajectory toward a low-scoring anchor, while the reverse process learns to progressively move trajectories toward higher driving scores conditioned on their current quality levels.
By supervising these ordered transitions with explicit quality scores, each reverse step is directly associated with measurable utility improvement.
Consequently, driving quality is embedded into the trajectory generation, guiding diffusion toward higher-quality trajectories rather than higher likelihood under the demonstration distribution.

Our contributions are summarized as follows:
\begin{itemize}
    \item We introduce quality-axis diffusion, which replaces Gaussian noise with driving quality as the diffusion coordinate, shifting generation from demonstration recovery to utility improvement.
    
    \item We propose QualiDrive, a framework that combines quality-conditioned trajectory refinement with multi-objective scoring for progressive generation and final selection.
    
    \item We achieve state-of-the-art performance on NAVSIM v1 and v2, reaching 92.74 EPDMS on v2---2.34 points above the previous best, while substantially increasing the density of high-quality trajectory candidates.
\end{itemize}
    
    

\section{Related Work}

\subsection{End-to-End Trajectory Generation}

End-to-end autonomous driving directly maps sensor observations to future
trajectories within a unified learnable system. Early planning-oriented
frameworks predict trajectories from structured scene representations or
planning queries~\citep{hu2023uniad,jiang2023vad,zeng2019nmp,casas2021mp3}. More recent proposal-based
planners construct multiple candidates from learned queries, anchors, or
factorized vocabularies, then select a final plan using scene-conditioned
scores~\citep{niu2025ipad,sun2026sparsedrivev2,ang2026clover,varadarajan2022multipath}. Hydra-MDP
distills human and rule-based teachers, while CLOVER learns evaluator
sub-scores for proposal ranking~\citep{li2024hydramdp,ang2026clover}. Online
trajectory evaluation and register-token scene compression further strengthen
candidate scoring~\citep{wote2025,kirby2026drivor}. These methods improve
coverage and selection, but generation remains bounded by learned proposal
sets or discrete trajectory vocabularies, motivating flexible generative
models.

\subsection{Diffusion-Based Trajectory Generation}

Diffusion provides a flexible way to model multimodal trajectory
distributions through iterative generation~\citep{chi2023diffusionpolicy}.
In end-to-end driving, DiffusionDrive combines multimodal trajectory anchors
with a truncated diffusion schedule for efficient and diverse
generation~\citep{liao2025diffusiondrive}, while DiffRefiner uses diffusion to
refine coarse proposals toward stronger scene compliance
~\citep{yin2026diffrefiner}. Subsequent work enhances diffusion planning with
additional optimization signals. Diffusion Planner uses classifier guidance
to incorporate safety constraints, goals, and driving styles
~\citep{dhariwal2021guided,zheng2025diffusionplanner}; SafeDiffuser imposes
safety constraints during the reverse process~\citep{xiao2023safediffuser};
and Diffusion-ES combines a diffusion prior with gradient-free evolutionary
search for black-box rewards~\citep{yang2024diffusiones}. DiffusionDriveV2
uses reinforcement learning to suppress low-quality modes and explore beyond
demonstrations~\citep{zou2025diffusiondrivev2}, while Gen-Drive improves
generation with a learned reward model and reinforcement-learning
fine-tuning~\citep{huang2024gendrive}. Despite their different mechanisms,
these approaches retain diffusion time or noise level as the generative
coordinate and introduce driving quality through guidance, constraints,
search, or reward optimization. Cold Diffusion shows that diffusion need not
be tied to Gaussian corruption, as paired degradation and restoration
operators can define a generative process~\citep{bansal2022cold}. Building on
this view, QualiDrive uses measured driving quality as the diffusion
coordinate, directly learning lower-to-higher-quality trajectory refinement.

\section{Method}

\begin{figure*}[t]
    \centering
    \includegraphics[width=0.9\textwidth]{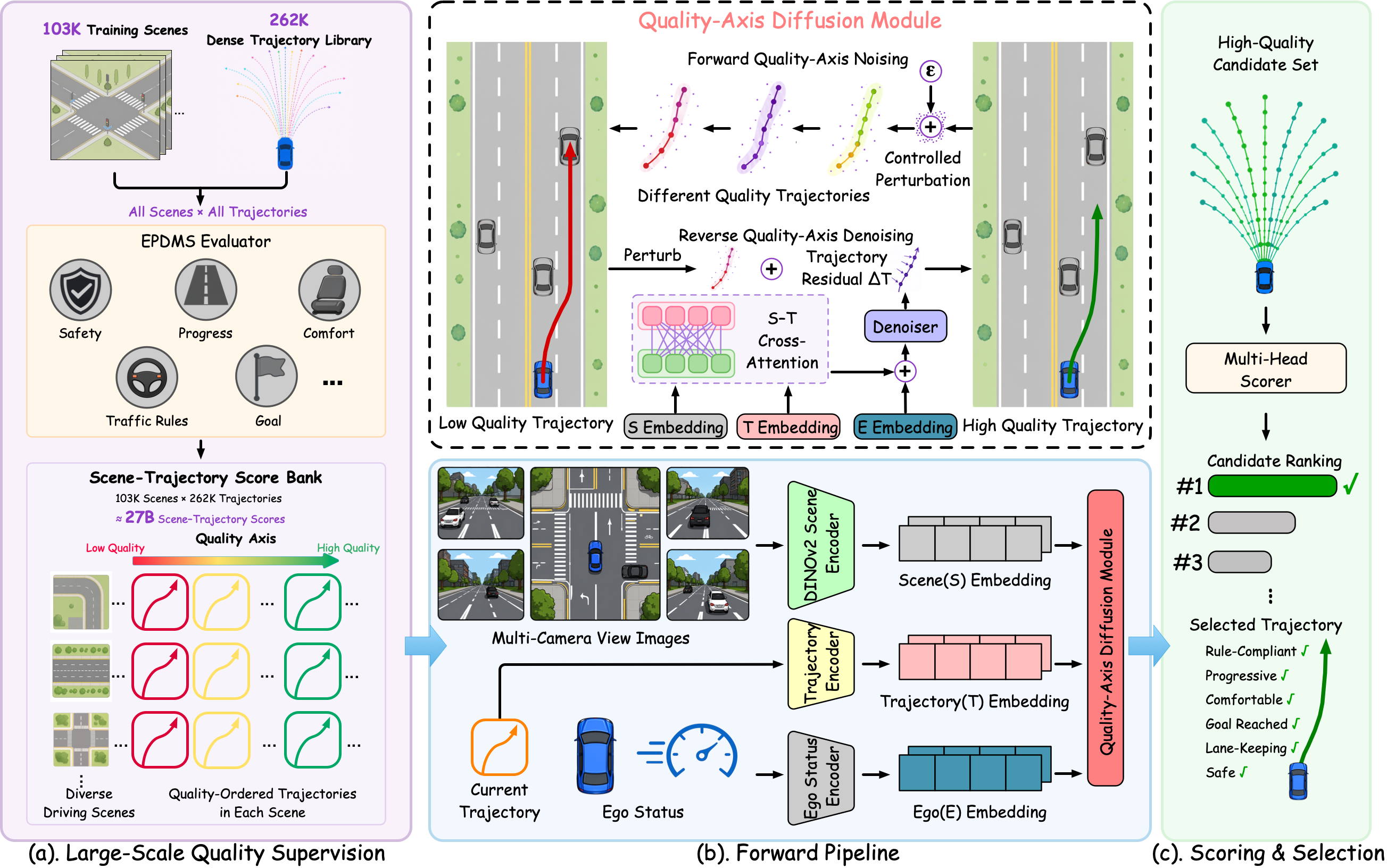}
    \caption{\textbf{Overview of Quality-Axis Diffusion.} (a) A dense
    scene--trajectory score bank orders same-intent trajectories by quality.
    (b) QualiDrive reverses this degradation through successive
    low-to-high-quality refinement. (c) A multi-objective scorer ranks the
    resulting candidates.}
    \label{fig:framework}
\end{figure*}

\subsection{Problem Formulation}

Let a driving scene be $s=(I,e)$, where $I$ contains the synchronized
multi-view images and $e$ is the current ego state. Given a planning horizon of
$H$ steps, the planner predicts a trajectory
$\tau\in\mathbb{R}^{H\times 3}$ containing planar position and heading. A
scene-conditioned quality function $Q(s,\tau)\in[0,1]$ evaluates the joint
satisfaction of safety, traffic rules, progress, lane keeping, and comfort.
Such multi-objective evaluation follows planning-oriented metrics developed
for non-reactive driving benchmarks
~\citep{dauner2023pdmclosed,dauner2024navsim}.
The objective is not merely to produce a likely trajectory under the logged
driving distribution, but to construct a candidate set
$\mathcal{C}(s)=\{\tau_k\}_{k=1}^{K}$ containing high-quality solutions and
select the strongest candidate under these objectives.

Standard diffusion planning adds Gaussian noise to a demonstrated trajectory
$\tau$ over timesteps $t$~\citep{ho2020ddpm,song2021scoresde}:
\begin{equation}
    x_t=\sqrt{\bar{\alpha}_t}\,\tau+
    \sqrt{1-\bar{\alpha}_t}\,\epsilon,
    \qquad \epsilon\sim\mathcal{N}(0,I),
    \label{eq:noise_axis}
\end{equation}
where $\bar{\alpha}_t$ controls noise strength and the reverse process recovers
$\tau$. Thus, $t$ describes signal-to-noise ratio rather than whether the
trajectory is safe, comfortable, or efficient. QualiDrive instead introduces a
quality coordinate $q$ and observed ordered states
$\tau^0,\ldots,\tau^L$ satisfying
$Q(s,\tau^{l+1})\geq Q(s,\tau^l)$. Its reverse process learns the transition
from $(\tau^l,q^l)$ to $(\tau^{l+1},q^{l+1})$. Diffusion time is thus replaced
by a semantic coordinate whose direction aligns with the planning objective.

\subsection{Large-Scale Quality Supervision}

\paragraph{Scene--trajectory score bank.}
Human demonstrations provide only one feasible trajectory per scene. We instead
use a dense library $\mathcal{V}=\{\tau_j\}_{j=1}^{M}$ of $M=262{,}144$ valid
trajectories and evaluate every member in each of the $N=103{,}288$ training
scenes, yielding
\begin{equation}
    B_{ij}=Q(s_i,\tau_j), \qquad B\in\mathbb{R}^{N\times M},
    \label{eq:score_bank}
\end{equation}
with approximately 27 billion scene--trajectory labels. $B_{ij}$ is scene
dependent: the same maneuver can be safe in one scene and unsafe in another.

\paragraph{Intent-preserving quality ladders.}
Sorting solely by score could connect unrelated behaviors, such as going
straight and turning. We therefore group the library into spatial path
prototypes with multiple motion profiles. For each scene and coarse intent, a
high-quality endpoint fixes the path prototype, and six observed variants of
that path form the ladder
\begin{equation}
    \mathcal{L}_{i,m}=\{(\tau_i^l,r_i^l)\}_{l=0}^{5}, \qquad
    Q(s_i,\tau_i^{l+1})\geq Q(s_i,\tau_i^l),
    \label{eq:quality_ladder}
\end{equation}
where $m$ indexes intent. $\tau_i^5$ is the highest-scoring valid variant, while
$\tau_i^0$ is a lower-quality variant near the corresponding intent anchor.
For intermediate level $l$, the target score is
\begin{equation}
    \bar{Q}_i^l=Q(s_i,\tau_i^0)+\frac{l}{5}
    \left[Q(s_i,\tau_i^5)-Q(s_i,\tau_i^0)\right]
    \label{eq:target_quality}
\end{equation}
We select a non-regressing library member that balances deviation from
$\bar{Q}_i^l$ and geometric change from the adjacent node. If no better variant
exists, the preceding node is carried forward; no trajectory is interpolated,
so every ladder state remains an observed member of $\mathcal{V}$.

The measured progress within each scene--intent ladder is
\begin{equation}
    r_i^l=\frac{Q(s_i,\tau_i^l)-Q(s_i,\tau_i^0)}
    {\max\{Q(s_i,\tau_i^5)-Q(s_i,\tau_i^0),\epsilon\}},
    \label{eq:quality_coordinate}
\end{equation}
with $r_i^0=0$, $r_i^5=1$, and $\epsilon=10^{-4}$. Here $r_i^l$ is relative
progress within a ladder, not an absolute EPDMS value. The level-wise training
means define the shared model schedule
$q=(0,.28,.50,.67,.88,1)$, which conditions the denoiser identically during
training and inference. Thus, the path group preserves maneuver identity,
$r_i^l$ grounds supervision in scene-dependent quality, and $q^l$ identifies
the common refinement stage.

\subsection{Quality-Axis Diffusion}

\paragraph{Forward quality degradation.}
Standard diffusion constructs intermediate states by adding progressively
stronger Gaussian noise to a clean sample. QualiDrive instead defines its forward
process by traversing an observed ladder in the opposite direction,
$\tau_i^5\rightarrow\cdots\rightarrow\tau_i^0$. Each level is a complete
trajectory already evaluated in the corresponding scene. On top of this,
we add a small Gaussian perturbation
$\widetilde{\tau}_i^l=\tau_i^l+\sigma_l\epsilon$,
$\epsilon\sim\mathcal{N}(0,I)$, for local robustness and diversity. Here $q^l$
specifies the shared semantic refinement stage, whereas $\sigma_l\epsilon$ only prevents
overfitting to the finite library. Every unperturbed $\tau_i^l$ is a valid,
scene-scored trajectory rather than an interpolation or noisy curve. At
inference, stochasticity is introduced only in the initial low-quality states;
no fresh noise is injected between quality transitions. The resulting sequence
traces each trajectory lineage through progressively higher-quality states.

\paragraph{Ordered local improvement.}
A direct low-to-high mapping must explain timing, lateral displacement, and
interaction changes in one update. The ladder instead decomposes them into
adjacent transitions with measured targets. Conditioning on
$(q^l,q^{l+1})$ distinguishes the requested improvement from current geometry
and supervises the shared denoiser at every stage. The five-step process thus
learns local, intent-preserving corrections rather than repeatedly applying an
endpoint regressor. Table~\ref{tab:generator_ablation} isolates this benefit
against Endpoint-Only Quality.

\paragraph{Quality-conditioned denoiser.}
The reverse process moves from $q^l$ to $q^{l+1}$. A DINOv2-based scene encoder
and a trajectory encoder produce features $F$ and $G$, which are fused by
cross-attention:
\begin{equation}
\begin{aligned}
    F&=\operatorname{SceneEncoder}(I),\\
    G&=\operatorname{TrajectoryEncoder}(\widetilde{\tau}^l),\\
    A&=\operatorname{LN}\!\left(G+\operatorname{Attn}(G,F,F)\right),
\end{aligned}
    \label{eq:feature_fusion}
\end{equation}
where $\operatorname{LN}$ denotes layer normalization and
$\operatorname{Attn}$ denotes cross-attention.
$f_\theta$ then takes $F$, $e$, $\widetilde{\tau}^l$, and $q^l,q^{l+1}$,
predicting a residual:
\begin{equation}
    \widehat{\tau}^{l+1}=\widetilde{\tau}^{l}+
    f_\theta(F,e,\widetilde{\tau}^{l},q^l,q^{l+1}).
    \label{eq:quality_update}
\end{equation}
This is repeated for $l=0$ through $l=4$, yielding the high-quality trajectory
$\widehat{\tau}^5$ after five transitions.

\subsection{Multi-Objective Scoring and Selection}

The generated set may contain several high-quality solutions that trade off
different objectives. We therefore predict nine scene-conditioned sub-scores
per candidate: no-at-fault collision ($\mathrm{NC}$), drivable-area
compliance ($\mathrm{DAC}$), driving-direction compliance ($\mathrm{DDC}$),
traffic-light compliance ($\mathrm{TL}$), time to collision ($\mathrm{TTC}$),
ego progress ($\mathrm{EP}$), lane keeping ($\mathrm{LK}$), history comfort
($\mathrm{HC}$), and extended comfort ($\mathrm{EC}$). The composite score is
\begin{equation}
\begin{split}
    \widehat{Q}_{\mathrm{sel}}={}&
    \widehat{m}_{\mathrm{NC}}\widehat{m}_{\mathrm{DAC}}
    \widehat{m}_{\mathrm{DDC}}\widehat{m}_{\mathrm{TL}}\\
    &\cdot\frac{5\widehat{m}_{\mathrm{TTC}}+5\widehat{m}_{\mathrm{EP}}
    +2\widehat{m}_{\mathrm{LK}}+2\widehat{m}_{\mathrm{HC}}
    +2\widehat{m}_{\mathrm{EC}}}{16},
    \label{eq:selection_score}
\end{split}
\end{equation}
The scorer is trained on candidate-level labels from the training score bank
with masked binary cross-entropy; an unavailable objective does not supervise
its head. At inference, it estimates these sub-scores from the scene and
candidate alone, without access to evaluator outputs, and serves only to rank
the generated set. The executed trajectory is
$\tau^*=\arg\max_{\tau\in\mathcal{C}}\widehat{Q}_{\mathrm{sel}}(s,\tau)$.

\subsection{Training and Inference}

Algorithm~\ref{alg:quality_axis} summarizes training and inference. The score
bank and ladders are training-only; inference uses the learned denoiser, quality
schedule, and multi-objective scorer.

\paragraph{Generator optimization.}
Training starts from perturbed intent anchors and unrolls the same five quality
transitions used at inference. The loss is
\begin{equation}
\begin{split}
    \mathcal{L}_{\mathrm{Gen}}={}&\sum_{l=0}^{4}w_l
    \ell_{\mathrm{traj}}(\widehat{\tau}^{l+1},\tau^{l+1})\\
    &+\lambda_s\sum_{l=0}^{4}w_l
    \ell_{\mathrm{smooth}}(\widehat{\tau}^{l+1},\tau^{l+1})\\
    &+\lambda_q\mathcal{L}_{\mathrm{quality}}
    +\lambda_r\mathcal{L}_{\mathrm{rank}}
    +\lambda_e\ell_{\mathrm{traj}}(\widehat{\tau}^{5},\tau^{5}),
    \label{eq:gen_loss}
\end{split}
\end{equation}
where $\ell_{\mathrm{traj}}$ and $\ell_{\mathrm{smooth}}$ are trajectory
regression and smoothness losses, $\mathcal{L}_{\mathrm{quality}}$ regresses the
predicted score to the measured ladder score, $\mathcal{L}_{\mathrm{rank}}$
enforces monotonic improvement, and the final term applies an additional
regression loss on the endpoint with weight $\lambda_e$.
\begin{equation}
    \mathcal{L}_{\mathrm{rank}}=\frac{1}{5}\sum_{l=0}^{4}
    \left[\gamma_l+\widehat{Q}(s,\widehat{\tau}^{l})-
    \widehat{Q}(s,\widehat{\tau}^{l+1})\right]_+,
    \label{eq:rank_loss}
\end{equation}
with $\gamma_l$ proportional to the measured score gap between adjacent levels.
The multi-objective scorer is trained separately on the same score bank and does
not contribute gradients to the generator during evaluation.

\paragraph{End-to-end inference.}
Initial perturbations around coarse intent anchors produce low-quality states.
Equation~\ref{eq:quality_update} is applied deterministically five times to
progressively refine each trajectory lineage without fresh noise. The generator
constructs $K=512$ final candidates for the multi-objective scorer.

\begin{algorithm}[tb]
\caption{QualiDrive: Learning and Sampling on the Quality Axis}
\label{alg:quality_axis}
\textbf{Input}: scenes $\{s_i\}$, score bank $B$, training library $\mathcal{V}$,
quality schedule $q^0<\cdots<q^5$, and candidate budget $K$
\begin{algorithmic}[1]
\STATE \textbf{Quality-axis training}
\FOR{each scene--intent pair $(s_i,m)$}
    \STATE Construct $\mathcal{L}_{i,m}$ from $B_{i,:}$ and same-intent
    members of $\mathcal{V}$ using Eqs.~\ref{eq:quality_ladder}--\ref{eq:quality_coordinate}.
\ENDFOR
\FOR{each sampled ladder $\mathcal{L}_{i,m}$}
    \STATE $F_i\leftarrow\operatorname{SceneEncoder}(I_i)$,
    $\widehat{\tau}_i^0\leftarrow\tau_i^0+\sigma_0\epsilon$,
    $\epsilon\sim\mathcal{N}(0,I)$
    \FOR{$l=0,\ldots,4$}
        \STATE $\widehat{\tau}_i^{l+1}\leftarrow\widehat{\tau}_i^l+
        f_\theta(F_i,e_i,\widehat{\tau}_i^l,q^l,q^{l+1})$
    \ENDFOR
    \STATE Update $\theta$ by minimizing $\mathcal{L}_{\mathrm{Gen}}$ in
    Eq.~\ref{eq:gen_loss}.
\ENDFOR
\STATE \textbf{Quality-axis inference}
\STATE $F\leftarrow\operatorname{SceneEncoder}(I)$,
$\mathcal{C}\leftarrow\varnothing$
\STATE Initialize trajectory lineages around coarse intent anchors
\FOR{each initialized lineage $g$}
    \FOR{$l=0,\ldots,4$}
        \STATE $\widehat{\tau}_g^{l+1}\leftarrow\widehat{\tau}_g^l+
        f_\theta(F,e,\widehat{\tau}_g^l,q^l,q^{l+1})$
    \ENDFOR
\ENDFOR
\STATE $\mathcal{C}\leftarrow
\operatorname{ConstructCandidatePool}(\{\widehat{\tau}_g^5\}_g,K)$,
$K=512$
\STATE \textbf{return} $\tau^*=\arg\max_{\tau\in\mathcal{C}}
\widehat{Q}_{\mathrm{sel}}(s,\tau)$ using Eq.~\ref{eq:selection_score}.
\end{algorithmic}
\end{algorithm}

\section{Experiments}

\begin{figure*}[t]
\savebox{\mainresultstablebox}{%
    \footnotesize
    \setlength{\tabcolsep}{0.45mm}%
    \begin{tabular}{l l c c c c c c c c c c c}
        \toprule
        Method & Backbone & NC & DAC & DDC & TL & EP & TTC & LK & HC & EC & PDMS & EPDMS \\
        \midrule
        Hydra-MDP++ & ResNet-34
            & 98.4 & 98.0 & 99.4 & 99.8 & 87.5 & 97.7 & 95.3 & 98.3 & 77.4 & 91.0 & 85.1$^{*}$ \\
        DiffusionDrive & ResNet-34
            & 98.2 & 96.2 & 99.5 & 99.8 & 87.4 & 97.4 & \textbf{97.0} & 98.3 & 87.8 & 88.1 & 84.8$^{*}$ \\
        WoTE & ResNet-34
            & 98.5 & 96.8 & 98.8 & 99.8 & 86.1 & 97.9 & 95.5 & 98.3 & 82.9 & 88.3 & 84.2 \\
        iPad & ResNet-34
            & 98.7 & 97.8 & 99.1 & 99.8 & 84.0 & 98.0 & 96.0 & 98.0 & 68.2 & 91.7 & 84.1 \\
        DriveSuprim & ResNet-34
            & 97.8 & 97.9 & 99.5 & \textbf{99.9} & \underline{90.6} & 97.1 & 96.6 & 98.3 & 77.9 & 93.5 & 86.0$^{*}$ \\
        DiffusionDriveV2 & ResNet-34
            & 97.7 & 96.6 & 99.2 & 99.8 & 88.9 & 97.2 & 96.0 & 97.8 & \underline{91.0} & 91.2 & 87.5 \\
        DrivoR & DINOv2-S
            & \underline{99.2} & 98.7 & \underline{99.6} & 99.8 & 79.8 & \underline{98.7} & 94.5 & 98.3 & 80.6 & \underline{93.7} & 88.4 \\
        SparseDriveV2 & ResNet-34
            & 98.1 & 98.1 & \underline{99.6} & 99.8 & \textbf{91.1} & 97.3 & \underline{96.9} & 98.2 & 78.4 & 92.0 & 90.1 \\
        CLOVER & DINOv2-S
            & \textbf{99.4} & \textbf{99.3} & 99.5 & 99.8 & 86.9 & \textbf{98.9} & 95.2 & 98.2 & 75.5 & \textbf{94.5} & \underline{90.4} \\
        \midrule
        \textbf{QualiDrive} & DINOv2-S
            & 98.4 & \underline{99.1} & \textbf{99.7} & \textbf{99.9} & 88.9 & 97.8 & 95.8 & 98.3 & \textbf{98.2} & \textbf{94.5} & \textbf{92.74} \\
        \bottomrule
    \end{tabular}}
\noindent
\begin{minipage}[t]{0.72\textwidth}
    \hrule height 0pt
    \centering
    \captionof{table}{Quantitative comparison on NAVSIM \texttt{navtest}. PDMS and
    EPDMS follow the v1 and v2 protocols, respectively; EPDMS is primary.
    The \textbf{best} and \underline{second-best} results are highlighted.
    $^{*}$ marks results reported under the original v2 evaluation protocol.}
    \label{tab:main_results}
    \usebox{\mainresultstablebox}
\end{minipage}\hfill
\begin{minipage}[t]{0.26\textwidth}
    \hrule height 0pt
    \centering
    \includegraphics[
        width=\linewidth,
        height=\dimexpr\ht\mainresultstablebox+\dp\mainresultstablebox\relax,
        keepaspectratio
    ]{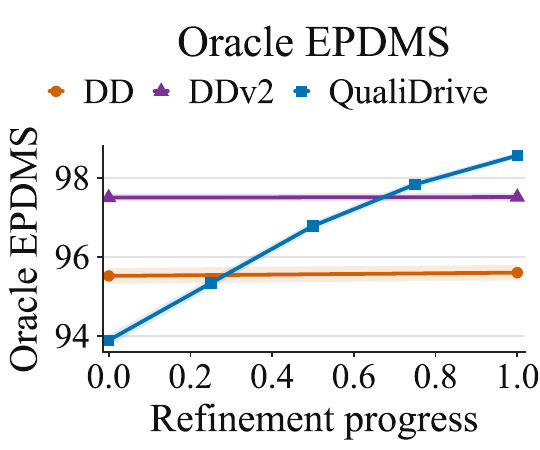}
    \captionof{figure}{Oracle EPDMS along the quality axis.}
    \label{fig:quality_evolution}
\end{minipage}
\end{figure*}

We evaluate QualiDrive in a progression from aggregate performance to mechanism
and, finally, qualitative behavior. We first compare complete planning
performance under the NAVSIM protocols, then separate candidate generation from
final selection and ablate the generator design. We next test whether the
quality axis produces consistent improvements across refinement stages and
scenes, before visualizing the resulting trajectories in representative cases.

\subsection{Benchmark and Evaluation Protocol}

We conduct experiments on NAVSIM~\citep{dauner2024navsim}, whose non-reactive
protocol builds on the Predictive Driver Model (PDM) family of planning metrics
~\citep{dauner2023pdmclosed}, evaluating on
\texttt{navtest} under both the v1 and v2 protocols. The primary metric is
EPDMS under NAVSIM v2, which aggregates nine complementary planning dimensions
We also report the Predictive Driver Model Score
(PDMS) under NAVSIM v1 for
compatibility with earlier work. QualiDrive uses a DINOv2-S
backbone~\citep{oquab2024dinov2} for visual encoding, and its
quality-axis generator produces trajectory candidates that are then ranked
by a multi-objective scorer. All evaluator-derived supervision is computed on
\texttt{navtrain}. On \texttt{navtest}, the official evaluator independently
scores the selected trajectory.

\subsection{Main Results}

Table~\ref{tab:main_results} compares QualiDrive with representative
end-to-end driving methods. Among methods using the same DINOv2-S backbone,
QualiDrive improves EPDMS from 88.4 for DrivoR and 90.4 for CLOVER to 92.74.
Beyond same-backbone settings, QualiDrive maintains a substantial margin over
earlier planning paradigms built upon traditional backbones~\citep{he2016resnet}. QualiDrive also
performs strongly across multiple EPDMS components, including
driving-direction compliance (DDC), traffic-light compliance (TL), hard
comfort (HC), and extended comfort (EC). On NAVSIM v1, QualiDrive achieves
94.5 PDMS, matching the best reported result. Performance under both protocols
supports the effectiveness of the quality axis as a diffusion coordinate.

\paragraph{Extended comfort.} EC is the official NAVSIM v2 two-frame
metric, scoring the consistency of a planner's successive selected
trajectories rather than the smoothness of any single one. Single-frame
comfort is already saturated near 98 for every method in
Table~\ref{tab:main_results}, so the EC column isolates frame-to-frame
consistency. QualiDrive treats this consistency as an explicit coordinate of
the quality axis and optimizes it jointly with the other objectives rather
than trading against them, attaining the highest reported EPDMS alongside its
EC.

\subsection{Candidate Generation and Selection}

\suppressfloats[t]

\begin{table}[t]
    \centering
    \caption{Candidate quality and final selection on NAVSIM v2.
    Counts are 20/800/200/64/512 by row; each uses its own selector.
    $^{*}$ follows Table~\ref{tab:main_results}.}
    \label{tab:candidate_quality}
    {\footnotesize
    \setlength{\tabcolsep}{0.7mm}
    \begin{tabular}{l c c c c c}
        \toprule
        & & \multicolumn{3}{c}{Candidate Quality} & \\
        \cmidrule(lr){3-5}
        Planner & Diversity & Oracle & Top-5 & Top-10 & EPDMS \\
        \midrule
        DiffusionDrive & 0.419 & 95.60 & 90.68 & 82.18 & 84.80$^{*}$ \\
        DiffusionDriveV2 & 0.387 & 97.51 & 96.85 & 96.37 & 87.50 \\
        SparseDriveV2 & 0.159 & 96.78 & 96.58 & 96.37 & 90.38 \\
        CLOVER & \textbf{0.487} & 97.81 & 97.06 & 96.21 & 90.40 \\
        \textbf{QualiDrive} & 0.362 & \textbf{98.57} & \textbf{98.21} & \textbf{97.89} & \textbf{92.74} \\
        \bottomrule
    \end{tabular}}
\end{table}

\begin{figure*}[t]
    \centering
    \begin{minipage}[t]{0.31\textwidth}
        \hrule height 0pt
        \centering
        \includegraphics[width=\linewidth]{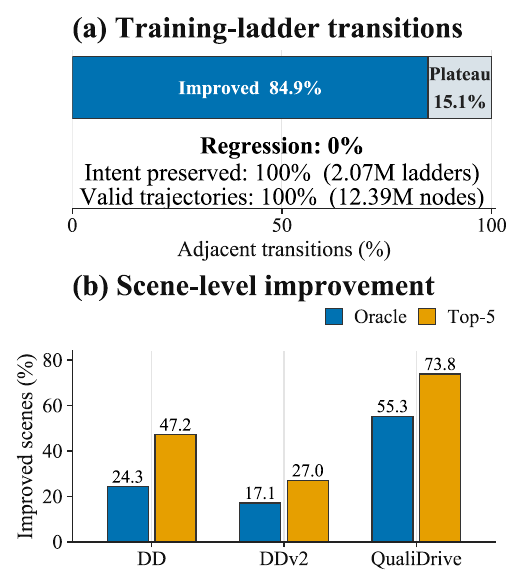}
    \end{minipage}\hfill
    \begin{minipage}[t]{0.67\textwidth}
        \hrule height 0pt
        \centering
        {\small\textbf{(c) Qualitative refinement}}\par
        \includegraphics[width=\linewidth]{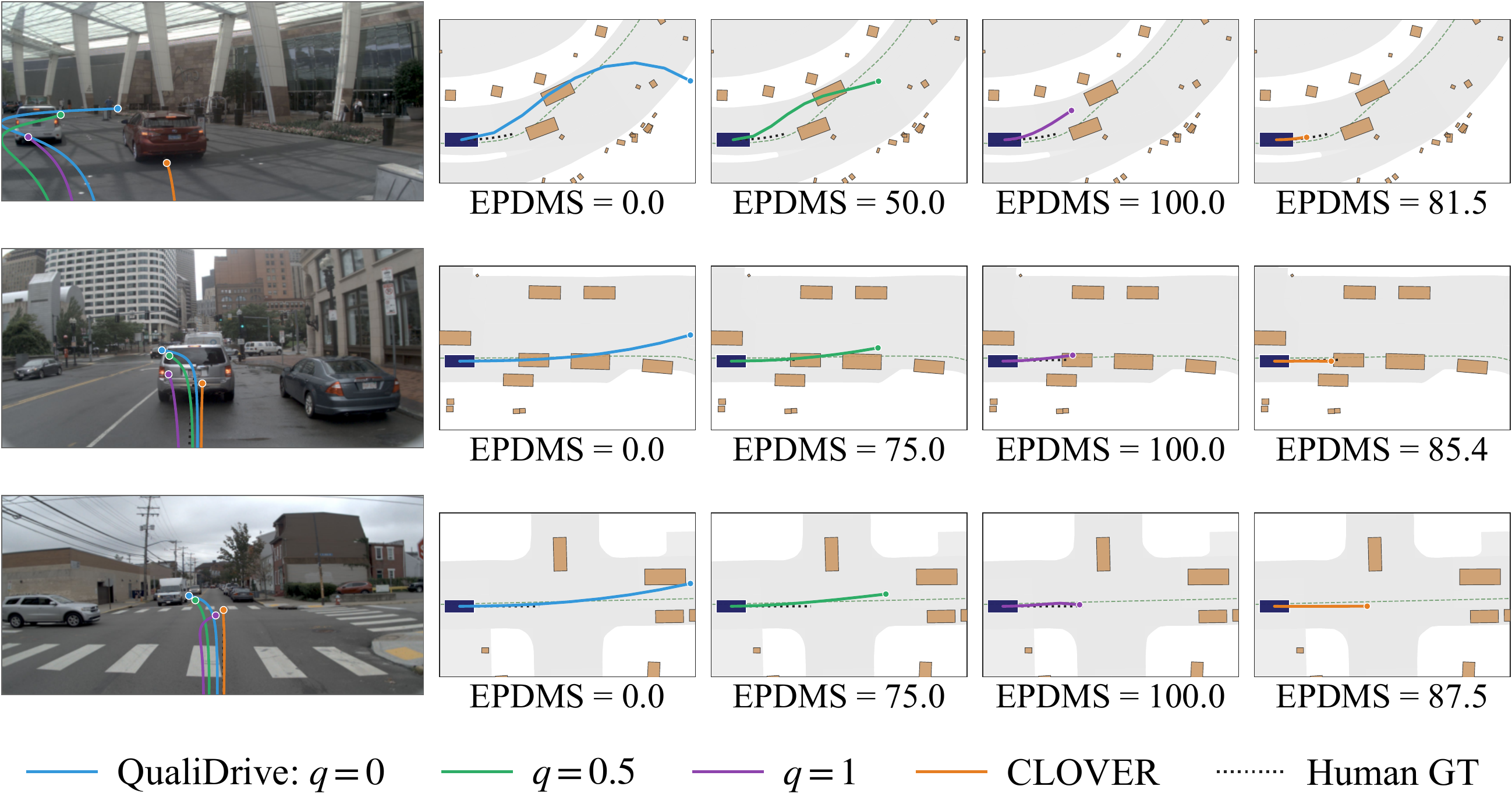}
    \end{minipage}
    \caption{\textbf{Quality-axis validity and qualitative refinement.}
    (a) Full-\texttt{navtrain} audit of 2.07 million six-level ladders.
    (b) Fraction of all 12,146 \texttt{navtest} scenes whose final valid pool
    improves over its first state. (c) QualiDrive refinement states versus
    CLOVER and human ground truth in representative scenes.}
    \label{fig:quality_evidence}
\end{figure*}

\paragraph{Candidate quality versus selected performance.}
Oracle, Top-5, and Top-10 measure the best, five-best, and ten-best true EPDMS
in each scene's candidate set, capturing both the upper bound and the density of
high-quality support. EPDMS evaluates each method's selected candidate.

Table~\ref{tab:candidate_quality} separates candidate-pool quality from final
selection. QualiDrive achieves the best Oracle, Top-5, Top-10, and selected
EPDMS at 98.57, 98.21, 97.89, and 92.74, respectively. These results indicate
that quality-aware generation and multi-objective scoring are complementary:
the generator exposes a strong set of feasible solutions, while the scorer
converts that support into the strongest final decision.
The three candidate statistics characterize different failure modes. Oracle
tests whether at least one strong solution is reachable, but can be dominated
by an isolated candidate that is difficult to identify. Top-5 and Top-10
average over progressively denser high-quality support and are therefore more
sensitive to whether generation repeatedly reaches useful solutions. The
simultaneous gains on all three metrics show that QualiDrive improves more than
the single best trajectory: it shifts the upper portion of the candidate pool
toward higher quality. The remaining gap between Oracle and selected EPDMS, on
the other hand, measures the difficulty faced by the learned selector. Reporting
both sides prevents a stronger scorer from being mistaken for a stronger
generator and makes their complementary contributions explicit.
Although CLOVER has the highest normalized pairwise diversity (0.487 versus
0.362), its Top-5 and Top-10 quality remains lower. Thus, geometric dispersion
alone does not explain QualiDrive's gains; the quality axis concentrates more
of its support in useful regions of the trajectory space.

\paragraph{Candidate-budget robustness.}
At $K=96$, QualiDrive achieves 98.01 Oracle and 91.80 selected EPDMS, already
above the strongest prior result of 90.40. Scaling to $K=512$ raises the
selected score to 92.74; doubling to $K=1024$ adds only 0.05 points at
$2.25\times$ the scoring time, supporting $K=512$ as the balanced operating
point.
At $K=512$, the complete candidate generator runs in 15.65 ms per scene at
batch size 16, while each selector model adds only 0.79--0.81 ms per scene.
Thus, the adopted budget combines dense high-quality support with practical
batched inference.

\begin{table}[t]
    \centering
    \caption{Effect of ordered multi-step quality supervision on NAVSIM v2
    ($K=512$). T5/T10 denote Top-5/Top-10 candidate EPDMS; Mean averages EPDMS
    over unique candidates. $Q\!\geq\!.95/.98$ are scene-averaged candidate
    percentages. Final uses the same five-model selector throughout.}
    \label{tab:generator_ablation}
    {\footnotesize
    \setlength{\tabcolsep}{0.2mm}
    \renewcommand{\arraystretch}{1.05}
    \begin{tabular}{
        >{\raggedright\arraybackslash}m{19mm}
        >{\centering\arraybackslash}m{8.5mm}
        >{\centering\arraybackslash}m{7.3mm}
        >{\centering\arraybackslash}m{7.3mm}
        >{\centering\arraybackslash}m{7.5mm}
        >{\centering\arraybackslash}m{11mm}
        >{\centering\arraybackslash}m{11mm}
        >{\centering\arraybackslash}m{8.5mm}}
        \toprule
        Generator & Oracle & T5 & T10 & Mean &
        $Q\!\geq\!.95$ & $Q\!\geq\!.98$ & Final \\
        \midrule
        \shortstack[l]{Standard\\Diffusion} & 97.75 & 97.25 & 96.86 & 37.64 & 10.64 & 7.46 & 91.31 \\
        \shortstack[l]{Endpoint-Only\\Quality} & 98.52 & 98.11 & 97.67 & 43.42 & 12.94 & 9.71 & 92.63 \\
        \textbf{QualiDrive} & \textbf{98.57} & \textbf{98.21} & \textbf{97.89} & \textbf{51.21} & \textbf{21.53} & \textbf{15.93} & \textbf{92.74} \\
        \bottomrule
    \end{tabular}}
\end{table}

\subsection{Generator Analysis}

\paragraph{Effect of ordered multi-step supervision.}
Table~\ref{tab:generator_ablation} compares all generators at the same $K=512$
candidate budget and with the same selector. Standard Diffusion provides
context for the effect of introducing quality supervision. Endpoint-Only
Quality then controls the incremental contribution of the ordered ladder: it
uses the same architecture and five generator updates but learns only the
direct relation between low-quality anchors and high-quality endpoints.
Relative to Endpoint-Only Quality, QualiDrive raises Oracle only modestly,
from 98.52 to 98.57. Its larger effect is on the density of strong candidates:
mean candidate EPDMS increases from 43.42 to 51.21, while the fractions above
0.95 and 0.98 EPDMS rise from 12.94\% and 9.71\% to 21.53\% and 15.93\%.
These are absolute gains of 8.59 and 6.22 percentage points, corresponding to
relative increases of 66.4\% and 64.1\%. Top-5 and Top-10 likewise improve
from 98.11/97.67 to 98.21/97.89. Thus, ordered multi-step supervision does not
merely uncover one slightly better Oracle trajectory; it makes high-quality
solutions substantially more frequent throughout the candidate pool.

The shared selector obtains 91.31, 92.63, and 92.74 EPDMS on the Standard
Diffusion, Endpoint-Only Quality, and QualiDrive pools, respectively.
Because the QualiDrive pool offers many more high-scoring alternatives, the
same selector also obtains a higher final score: 0.11 points above
Endpoint-Only Quality and 1.43 points above Standard Diffusion. The principal
benefit of the ordered quality axis is therefore denser high-quality support,
which makes strong candidates easier for the downstream selector to identify.

\paragraph{Progressive quality improvement along the generation path.}
Figure~\ref{fig:quality_evolution} shows the Oracle EPDMS along the generation
path for DD, DDv2, and QualiDrive. DD and DDv2 exhibit almost no improvement
across their two output states (gains of
only 0.08 and 0.01 points), with DDv2's gain
statistically indistinguishable from zero. This indicates that, for these
planners, standard diffusion essentially saturates at the first denoising step,
with the second step providing no meaningful improvement. In contrast,
QualiDrive increases monotonically across five
quality-axis states, from 93.89 to 98.57, a total gain of 4.68 points.
The intermediate states are not merely assigned larger coordinates---their
candidate sets have measurably higher and denser quality distributions.

\subsection{Quality-Axis Validity}

Figure~\ref{fig:quality_evidence}(a) audits the supervision before model
training. Across 2,065,760 intent-conditioned ladders and 12,394,560 observed
trajectory nodes, every adjacent transition is non-regressing and 84.9\% is
strictly improving. Every node is a complete valid trajectory, and no ladder
switches its spatial intent. These results verify that the coordinate is
grounded in measured scene-dependent quality rather than assigned to arbitrary
interpolants. The remaining 15.1\% of adjacent pairs are exact plateau carries,
not regressions, while the mean quality gain from the first to final ladder
node is 32.76 points; only 0.60\% of scene--intent pairs are degenerate.

Figure~\ref{fig:quality_evidence}(b) further tests whether the aggregate
gains in Figure~\ref{fig:quality_evolution} extend across scenes. From the first
to final valid state, QualiDrive raises scene-wise Oracle quality in 55.3\% of
navtest scenes and Top-5 quality in 73.8\%. The corresponding rates are only
24.3\%/47.2\% for DiffusionDrive and 17.1\%/27.0\% for DiffusionDriveV2. The
quality-axis improvement is therefore broadly distributed rather than being
driven by a small number of unusually large gains. Moreover, Top-5 quality
strictly increases at each successive transition in 58.3\%, 56.7\%, 56.5\%,
and 51.4\% of scenes, respectively. Every stage therefore contributes positive
coverage for a majority of scenes, even as the pool approaches saturation. The
cumulative fraction of scenes with an improved Oracle likewise grows from
36.5\% after the first transition to 45.8\%, 51.3\%, and finally 55.3\%. The
95\% paired-bootstrap interval for the final 4.68-point Oracle gain is
$[4.55,4.81]$.

\paragraph{Quality-coordinate calibration.}
Table~\ref{tab:quality_coordinate_ablation} tests the shared calibrated
coordinate against directly conditioning on each ladder's measured progress.
At $K=96$ on a 512-scene validation subset, the shared coordinate improves
Oracle from 97.40 to 97.62, Top-5 from 95.63 to 96.11, and Top-10 from 93.49
to 94.61. A common stage coordinate therefore gives the shared denoiser a more
consistent refinement signal, while measured per-ladder progress remains the
basis for constructing ordered supervision.

\begin{table}[tb]
    \centering
    \caption{Quality-coordinate calibration at $K=96$ on a 512-scene
    validation subset.}
    \label{tab:quality_coordinate_ablation}
    \footnotesize
    \setlength{\tabcolsep}{2.5mm}
    \begin{tabular}{lccc}
        \toprule
        Conditioning & Oracle & Top-5 & Top-10 \\
        \midrule
        Per-ladder measured $r_i^l$ & 97.40 & 95.63 & 93.49 \\
        Shared calibrated $q^l$ & \textbf{97.62} & \textbf{96.11} & \textbf{94.61} \\
        \bottomrule
    \end{tabular}
\end{table}

\subsection{Qualitative Studies}

As shown in Figure~\ref{fig:quality_evidence}(c), QualiDrive progressively
improves trajectories along the quality axis across three challenging scenes.
In the left-turn scenario, the initial trajectory deviates well beyond the
road boundary; the intermediate state returns to the road but still risks
collision with a leading vehicle; the final trajectory generates a left-forward
path, reaching an EPDMS of 100. By contrast, the human ground-truth (GT)
trajectory conservatively follows the red car ahead, and CLOVER adopts an even more cautious strategy,
achieving only 81.5. In the urban straight-road scenario, QualiDrive's final
trajectory shifts slightly leftward rather than following the straight path
taken by both the human and CLOVER, substantially exceeding CLOVER's 85.4.
In the intersection scenario, QualiDrive adds a slight rightward deviation to
an otherwise conservative straight path, evading an oncoming vehicle from the
left. Both CLOVER and the human demonstration proceed straight, incurring
collision risk and scoring 87.5, clearly below QualiDrive. These cases show
that QualiDrive can move beyond imitation and surpass logged human trajectories
in benchmark-assessed quality.

\section{Conclusion}

QualiDrive replaces the Gaussian-noise coordinate with measured driving
quality and learns ordered refinement toward higher utility. It reaches 94.5
PDMS under NAVSIM v1 and 92.74 EPDMS under v2. At matched $K=512$, its Oracle
gain over Endpoint-Only Quality is small (98.57 versus 98.52), but the
candidate fractions above 0.95 and 0.98 rise by 66.4\% and 64.1\%, raising the
same selector from 92.63 to 92.74. Thus, quality-axis supervision makes strong
trajectories more prevalent and easier to select rather than merely finding
one exceptional candidate. Methodologically, QualiDrive requires a large
offline score bank built from a finite trajectory library and five sequential
refinement updates, increasing supervision and inference cost. Evaluation is
also limited to non-reactive NAVSIM and evaluator-derived supervision;
reactive closed-loop behavior, independent safety or preference judgments, and
out-of-distribution generalization remain important future tests.

\bibliographystyle{unsrtnat}
\bibliography{aaai2027}

\end{document}